\documentclass{article}
\usepackage{graphicx} 
\PassOptionsToPackage{numbers, compress, sort}{natbib}
\usepackage[preprint]{neurips_2024}
\usepackage{amsmath, amssymb, amsfonts}
\usepackage{times}
\usepackage{bm}
\usepackage{url} 
\usepackage{hyperref}
\usepackage{cleveref}
\usepackage{booktabs}
\usepackage{makecell}
\usepackage{multirow}
\usepackage{color, colortbl}
\usepackage{xcolor}
\usepackage{subcaption}
\usepackage{booktabs}
\usepackage{multirow}
\usepackage{tabularx}
\usepackage{epigraph}
\usepackage{caption, multicol}
\hypersetup{
            colorlinks=true,                
            breaklinks=true,                
            urlcolor= black,                
            linkcolor= blue,                
            citecolor=blue,
            filecolor=black,
            linkbordercolor=blue
            }
\usepackage{soul}
\usepackage{tabularx}

\newcommand{\christos}[2]{\textcolor{red}{\st{#1}} \textcolor{green}{#2}}

\title{AllMem: A Memory-centric Recipe for Efficient Long-context Modeling}

\author{\textbf{Ziming Wang$^{\star,1,3}$, Xiang Wang$^{\star,1,3}$, Kailong Peng$^1$, Lang Qin$^1$,} \\
  \textbf{Juan Gabriel Kostelec$^2$, Christos Sourmpis$^2$, Axel Laborieux$^2$, Qinghai Guo$^{\dagger,1,3}$} \\
  $^1$ ACS Lab, Huawei Technologies \quad $^2$ Huawei Switzerland\\
  $^3$ \{wangziming33, wangxiang224, guoqinghai\}@huawei.com \\
  $\star$ Equal Contribution
}

\date{February 2026}

\begin{document}

\maketitle

\begin{abstract}
Large Language Models (LLMs) encounter significant performance bottlenecks in long-sequence tasks due to the computational complexity and memory overhead inherent in the self-attention mechanism. To address these challenges, we introduce \textsc{AllMem}, a novel and efficient hybrid architecture that integrates Sliding Window Attention (SWA) with non-linear Test-Time Training (TTT) memory networks. \textsc{AllMem} enables models to effectively scale to ultra-long contexts while mitigating catastrophic forgetting. This approach not only overcomes the representation constraints typical of linear memory models but also significantly reduces the computational and memory footprint during long-sequence inference. Furthermore, we {implement} a Memory-Efficient Fine-Tuning strategy to replace standard attention layers in pre-trained models with memory-augmented sliding window layers. This framework facilitates the efficient transformation of any off-the-shelf pre-trained LLM into an \textsc{AllMem}-based architecture. Empirical evaluations confirm that our 4k window model achieves near-lossless performance on 37k LongBench with a marginal 0.83 drop compared to full attention. Furthermore, on InfiniteBench at a 128k context, our 8k window variant outperforms full attention, which validates the effectiveness of our parameterized memory in mitigating noise and maintaining robust long-range modeling without the prohibitive costs of global attention.
\end{abstract}



\epigraph{"Memory is the residue of thought."}{--- Daniel T. Willingham}

\section{Introduction}
In long-sequence language modeling, the decoder-only Transformer architecture \cite{vaswani2017attention} remains dominant, largely due to the proficiency of the self-attention mechanism in capturing global dependencies \cite{bahdanau2014neural}. However, this architecture faces significant bottlenecks: it fails to effectively model sequences exceeding its pre-trained context window, and the use of Softmax attention incurs prohibitive computational and storage costs that scale with the input sequence length $L$ \cite{gu2024mamba}. On edge devices, such as smartphones, the $O(L^2)$ computational complexity results in a substantial spike in inference latency, severely degrading the user experience for long-context tasks. Furthermore, the $O(L)$ storage complexity of the KV Cache makes it impossible to maintain the entire context within limited memory constraints{. These challenges arise not only on edge devices, but also in server-grade systems, as recent trends in agents and thinking models produce an ever-increasing amount of tokens, making inference challenging even with abundant computational resources.}


To address the $O(L^2)$ computational bottleneck and the $O(L)$ memory access overhead of Softmax attention, researchers have proposed various{ mitigation strategies. For example, }linear attention methods \cite{dao2024transformers, yang2023gated, yang2024gated,team2025kimi} support parallelization during training and can be executed recurrently during inference {with a constant memory footprint and per-token computation}, significantly reducing computational and memory pressure for long sequences. {Another approach}, ``Test-Time Training'' (TTT) or ``Test-Time Memory'' \cite{sun2024learning, behrouz2024titans,zhang2025test,behrouz2025nested,vonoswald2025mesanetsequencemodelinglocally} treats sequence modeling as an online compression problem, introducing non-linear memory networks to facilitate information compression and establish a persistent long-term memory. 
{Finally, sparse} attention mechanisms such as sliding window attention with "sink tokens" \cite{xiao2023efficient} can drastically reduce the computational load of the attention matrix while preserving Softmax similarity metrics and the precise information of key tokens. However, these efficient attention mechanisms {often suffer from} information loss {leading to a reduction in} language modeling performance. 
To strike a balance between model performance and computational efficiency, recent research has focused on hybrid attention architectures \cite{lieber2024jamba,gu2025jet,junxiongdaniele2024mambainllama}. These designs aim to leverage efficient mechanisms to boost throughput while retaining the robust semantic modeling capabilities of standard attention.

Currently, mainstream hybrid architectures primarily manifest in two forms: inter-layer hybrids and intra-layer hybrids. Specifically, inter-layer hybrids \cite{qwen2025qwen3next, li2025minimax, team2025kimi, agarwal2025gpt} involve sequentially stacking different types of attention mechanisms across layers. This approach has become a prevalent choice for major LLM developers in their exploration of large-scale models, as exemplified by the GPT-OSS-120B \cite{agarwal2025gpt}, Qwen3-Next-80B-A3B \cite{qwen3next}, Ring-2.5-1T \cite{hfRing251T}, etc. Despite their engineering feasibility, these structures still rely on partial global attention mechanisms. Consequently, their computational complexity remains {$O(L^2)$}, which imposes a bound on potential efficiency gains. On the other hand, intra-layer hybrid architectures (see \cite{zuo2025falcon, dong2024hymba, behrouz2024titans}) aim to achieve linear computational complexity by fusing multiple attention mechanisms within a single Transformer layer. However, these approaches typically require training from scratch with specific post-training policies, incurring significant computational costs for exploring new architectures. The gap in downstream performance between the current intra-layer hybrid designs and the full-attention Transformer also limits their applications in real-world scenarios. 

{In this study, we aim to bridge the performance gap between efficient attention models and full attention models in long-context tasks. Our proposed \textsc{AllMem} model integrates a short-term ``all-seeing'' sliding window attention with a novel long-term memory mechanism. By facilitating efficient knowledge compression, \textsc{AllMem} reduces the computational complexity of Transformers to constant overhead. We evaluate the model on a series of long-context benchmarks, comparing pre-trained small Qwen3 models (0.6B and 1.7B) with fine-tuned Qwen3-\textsc{AllMem} variants. Results show that our model achieves or exceeds the performance of the baseline Qwen3 models while using significantly lower computational and memory resources—sometimes as little as 11\% of the original model.}

\section{Preliminaries}
From a memory-augmented perspective, the autoregressive sequence modeling can be fundamentally characterized as a reconstruction process of key-value pairs. The model aims to construct a mapping function $\phi_m: \mathbf{q}_t \to \hat{\mathbf{v}}_t$ using the historical collection $\{(\mathbf{k}_i, \mathbf{v}_i)\}_{i\le t}$, with the objective of minimizing the reconstruction loss $\mathcal{\ell}(\hat{\mathbf{v}}_t, \mathbf{v}_t)$. For the standard attention mechanism, the computation is formalized as:
$$
\hat{\mathbf{v}}_t = \phi_m(\mathbf{q}_t, \{(\mathbf{k}_i, \mathbf{v}_i)\}_{i\le t}) = \underset{\mathbf{v}}{\arg\min} \sum_{i=1}^t s(\mathbf{k}_i, \mathbf{q}_t) \|\mathbf{v}_i - \mathbf{v}\|_2^2 = \sum_{i=1}^t \frac{s(\mathbf{k}_i, \mathbf{q}_t)}{\sum_{j=1}^t s(\mathbf{k}_j, \mathbf{q}_t)} \mathbf{v}_i
$$
where $s(\cdot)$ measures the scaled dot-product similarities between query $\mathbf{q}$ and key $\mathbf{k}$. Statistically, attention is essentially a Nadaraya-Watson kernel regression (a non-parametric solution) for autoregressive tasks under similarity-weighted $L_2$ loss \cite{fan2018local}. While this non-parametric approach grants the model precise sequence backtracking capabilities, it necessitates maintaining a KV cache that grows linearly with sequence length and incurs {$O(L^2)$} computational complexity. This not only constrains inference efficiency on edge devices but also lacks higher-order memory processing mechanisms from a neuroscientific standpoint, such as memory consolidation, multi-timescale forgetting, and reverberation \cite{ji2007coordinated,kirkpatrick2017overcoming}.

To overcome these bottlenecks, linear attention models, exemplified by Mamba-2 \cite{dao2024transformers}, abandon the computationally intensive Softmax quadratic structure in favor of parallelizable linear recurrent State Space Models (SSMs). By simplifying the selective scan mechanism, the attention mechanism can be expressed as a linear recurrence in matrix form:
$$\mathbf{M}_t = \alpha_t \mathbf{M}_{t-1} + \mathbf{v}_t \mathbf{k}_t^\top \label{eq.mamba}$$
Alternatively, more sophisticated directional update rules, such as the Delta Rule \cite{yang2024gated}, can be introduced:
$$
\mathbf{M}_t = \mathbf{M}_{t-1} \left( \alpha_t \mathbf{I} - \beta_t \mathbf{k}_t \mathbf{k}_t^\top \right) + \beta_t \mathbf{v}_t \mathbf{k}_t^\top  \label{eq.delta_net}
$$
In this framework, $\mathbf{M}_t$ serves as a parametric linear memory unit storing the outer-product associative representations of historical KV pairs. Test-Time Training (TTT) \cite{sun2020test} provides a unified theoretical explanation for this process from the perspective of online learning: linear attention is effectively a parametric solution to the sequence modeling optimization problem under specific linear constraints. During sequence processing, the model utilizes the parametric memory unit $\mathcal{M}$ to compress historical information online. During the prediction phase, information is extracted from the memory unit based on the current query $\mathbf{q}_t$.

$$\hat{\mathbf{v}}_t = f(\mathcal{M}_t, \mathbf{q}_t) = \mathcal{M}_t (\mathbf{q}_t)$$
This procedure no longer relies on an explicit, fine-grained KV cache. For memory updates, TTT proposes a general design principle: the memory unit $\mathcal{M}_t$ is optimized online during inference via gradient descent to minimize the instantaneous reconstruction loss:
$$\mathcal{M}_t = \mathcal{M}_{t-1} - \eta_t \nabla_{\mathcal{M}_{t-1}} \mathcal{L}(f(\mathcal{M}_{t-1}, \mathbf{k}_t), \mathbf{v}_t)$$
Different model architectures correspond to distinct loss functions, which in turn derive various memory update rules. For instance, in Mamba-2, the corresponding loss function is represented as:$$\ell(\hat{\mathbf{v}}_t, \mathbf{v}_t) = -\mathbf{v}_t^\top \hat{\mathbf{v}}_t + \frac{1-\alpha_t}{2} \|\mathbf{M}_t\|_F^2 = -\mathbf{v}_t^\top \mathbf{M}_t \mathbf{k}_t + \frac{1-\alpha_t}{2} \|\mathbf{M}_t\|_F^2$$
In this work, we explore a novel architecture that augments sliding-window attention with test-time non-linear parametric memory. This framework facilitates the efficient inheritance of knowledge from pre-trained models while delivering superior performance and efficiency in long-context modeling.

\section{Method}
\subsection{Design Principles}
\label{subsec:design_principles}
Motivated by the limitations of existing memory mechanisms and inspired {from cognitive neuroscience}, {we summarize the philosophy of our model with the two following core ideas:}

\begin{enumerate}
    \item \textbf{Memory as {a Learning system}}: Memory should not be viewed merely as passive storage of information. Instead, it should function as an active, hierarchical learning system capable of abstracting, compressing, and {retrieving memories} knowledge. Effective long-term memory compression, therefore, arises not from raw token retention, but from the network's capacity to learn and distill {information}.

    \item \textbf{Learning Across Multiscale Temporal Dynamics}: Memory learning should be inherently multiscale, {treating} long-term and short-term {memory differently}. Long-{term} learning establishes stable, high-level reasoning patterns and persistent logical knowledge, while short-{term} learning enables fine-grained, context-sensitive adaptation and rapid state updates. These complementary temporal modes {should be} coupled through a shared representational space, enabling bidirectional interaction and dynamic mode switching.
\end{enumerate}

Building upon these principles, we propose a novel native parameterized memory architecture—\textsc{AllMem} (A Large Language MEMory model)—that integrates a hybrid design combining sliding-window attention (SWA) with a nonlinear, test-time trainable (TTT) memory network.
{Thanks to the modularity of our proposed memory model, we can efficiently fine-tune pretrained large language models to benefit from an adaptive multi-scale memory system.}

\subsection{AllMem}
\label{subsec:arch}

\begin{figure}[htb]
    \centering
    \includegraphics[width=\linewidth]{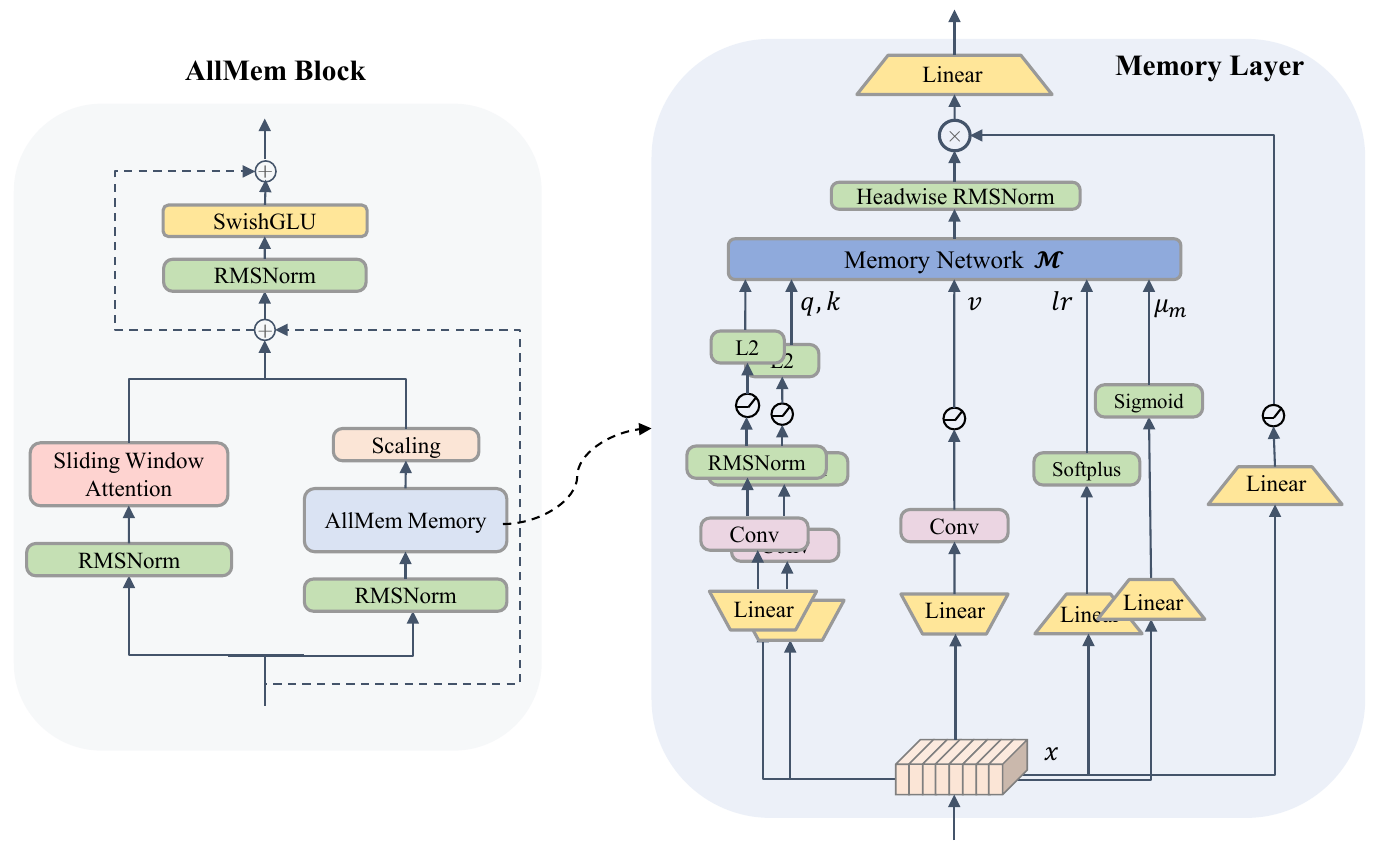}
    \caption[\christos{Overall}{Model} architecture of the memory module]{\textbf{{Model} architecture of the memory module}: (Left) A single decoder layer structure, where the \texttt{Token Mixer} consists of two parallel branches: a sliding-window attention (SWA) module for modeling local, fine-grained dependencies, and a long-term \textsc{AllMem} memory unit for capturing global, persistent semantic patterns. (Right) The internal meta-parameter structure of the TTT-enabled memory unit, including learnable momentum decay rates, learning rates, output gating branches, and QKV projection weights.}
    \label{fig.arch}
\end{figure}


Following the design principles, we introduce \textsc{AllMem}, a learnable, native parameterized memory fusion architecture. \textsc{AllMem} preserves strong performance on short-sequence tasks while significantly enhancing long-sequence modeling capacity, {leveraging a computationally efficient long-term memory strcture.}

The core innovation of \textsc{AllMem} lies in its parallel, modular architecture that explicitly decouples long-term persistent memory from {precise short-term} memory. These two modalities are dynamically fused via a learnable, channel-wise scaling gate mechanism, enabling adaptive, context-sensitive integration of memory contributions.

As illustrated in \cref{fig.arch}, the \textsc{AllMem} model adopts the standard Transformer decoder layer structure. The \texttt{Channel Mixer} reuses the pre-trained MLP weights from the base model, thereby preserving foundational semantic representation capabilities. The \texttt{Token Mixer}, {on the other hand, is constructed} as a parallel composition of two components:

\begin{itemize}
    \item A \textit{sliding-window attention }(SWA) module, which computes attention over a fixed-size local window of tokens, enabling efficient modeling of fine-grained local dependencies.
    \item A \textit{nonlinear, TTT-enabled memory network} (\textsc{AllMem}), which maintains and updates a long-term, compressed memory state through test-time online learning.
\end{itemize}

This parallel design achieves a significant reduction in computational complexity: Prefill complexity is reduced to $\mathcal{O}(LW)$, and Decode complexity to $\mathcal{O}(W)$, where $L$ is the input sequence length and $W$ is the sliding window size. To enable dynamic, context-aware fusion of the two memory modalities, we introduce a learnable, per-channel scaling coefficient $\bm{\alpha} \in \mathbb{R}^{d_{\text{model}}}$. The fusion is formalized as:

\[
\text{TokenMixer}(\mathbf{x}) = \text{SWA}\left(\text{RMSNorm}(\mathbf{x})\right) + \bm{\alpha} \cdot \textrm{\textsc{AllMem}}\left(\text{RMSNorm}(\mathbf{x}))\right)
\]

Here, $\bm{\alpha}$ is initialized to zero {vector}{}, allowing the model to start with a bias toward SWA. During training, the network gradually learns the optimal balance between local attention and long-term memory contributions per channel, stabilizing training and accelerating convergence.

During training, we freeze the weights of the SWA and Channel Mixer components, and only fine-tune the meta-parameters of the \textsc{AllMem} memory unit--specifically, the QKV projection matrices, the learnable learning rate and momentum decay coefficients, the RMSNorm parameters, the output gating branch, and the output projection layer. This selective fine-tuning strategy mitigates the risk of catastrophic forgetting,  preserving the model’s prior knowledge while enabling adaptive memory learning.

A critical challenge in such memory systems lies in achieving efficient {and} dynamic memory updates and compression. {I}nspired by the test-time training (TTT) paradigm, we adopt an online optimization strategy that continuously updates the Memory Network {($\mathcal{M}$)} parameters by minimizing a reconstruction loss {($\mathcal{L}$)}. The update rule is formulated as:
\begin{gather*}
        \mathcal{M}_t = \mathcal{M}_{t-1} - \eta_t \nabla_{\mathcal{M}_{t-1}} \mathcal{L}(f(\mathcal{M}_{t-1}, \mathbf{k}_t), \mathbf{v}_t) \\
    \mathcal{L}(f(\mathcal{M}_{t-1}, \mathbf{k}_t), \mathbf{v}_t)  = -\text{dot\_product} (f(\mathcal{M}_{t-1}, \mathbf{k}_t), \mathbf{v}_t)
\end{gather*}

\begin{figure}[t]
    \centering
    \includegraphics[width=\linewidth]{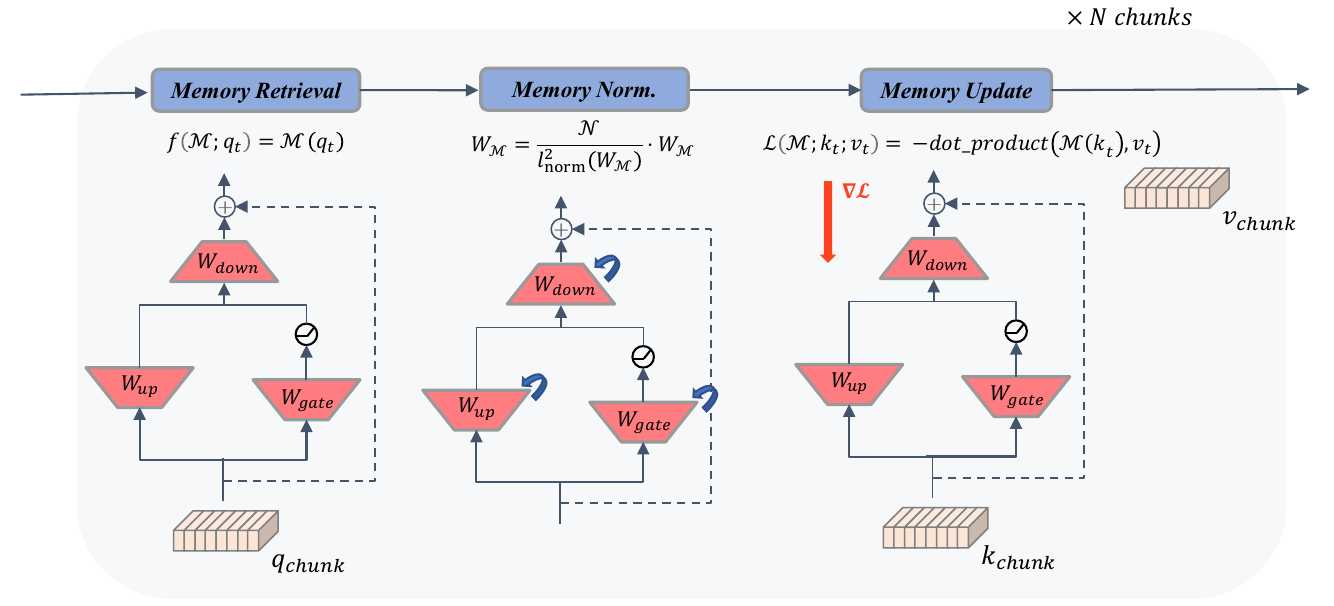}
    \caption[Online learning pipeline of the memory module]{\textbf{Online learning pipeline of the memory module}: To prevent future information leakage, we employ a strictly ordered sequence of operations--memory read, weight normalization, and memory update--ensuring stable and robust TTT dynamics.}
    \label{fig.pipeline}
\end{figure}

Unlike prior approaches such as AHN~\cite{fang2025artificial} and Mamba~\cite{gu2024mamba}, which rely on linear attention or linear memory matrices, our framework employs a residual-connected SwishGLU-based nonlinear memory unit $f(\cdot)$, enabling richer knowledge compression and context-aware representation learning. To ensure stable gradient flow and high parallelism during test-time training, we adopt a chunk-wise retrieval and update strategy with a relatively large chunk size $c$ , inspired by LaCT~\cite{zhang2025test}. The memory read operation $\phi_m: q_t \to \hat{v}_t$ is formulated as:
$$
\begin{aligned}
&\hat{\mathbf{V}}_{j} = \mathbf{A} \mathbf{W}_{\text{down}}^\top +\mathbf{Q}_{j} \\
&\mathbf{A} = \text{SiLU}(\mathbf{H}_{gate}) \odot \mathbf{H}_{in} \\
\mathbf{H}_{gate} &= \mathbf{Q}_{j} \mathbf{W}_{\text{gate}}^\top  \quad \quad \mathbf{H}_{in} = \mathbf{Q}_{j} \mathbf{W}_{\text{up}}^\top 
\end{aligned}
$$
where $\mathbf{Q}_{j} = [\mathbf{q}_{jc+1}, \mathbf{q}_{jc+2},..., \mathbf{q}_{jc+c}]^\top$ and $\hat{\mathbf{V}}_{j} = [\hat{\mathbf{v}}_{jc+1}, \hat{\mathbf{v}}_{jc+2},..., \hat{\mathbf{v}}_{jc+c}]^\top$ denote the queries and predictions for the $j$-th block, respectively. 
The corresponding test-time gradient computation is given by:
$$
    \begin{aligned}
\mathbf{G}_{out} &\triangleq \nabla_{\hat{\mathbf{V}}} \mathcal{L} = -\mathbf{V} \\
\mathbf{G}_{A} &\triangleq \nabla_{\mathbf{A}} \mathcal{L} = \mathbf{G}_{out} \mathbf{W}_\text{down} = -\mathbf{V} \mathbf{W}_\text{down} \\
\mathbf{G}_{in} &= \mathbf{G}_{A} \odot \text{SiLU}(\mathbf{H}_{gate}) \\
\mathbf{G}_{gate} &= \mathbf{G}_{A} \odot \mathbf{H}_{in} \odot \text{SiLU}'(\mathbf{H}_{gate}) \\
 \nabla_{\mathbf{W}_\text{gate}} \mathcal{L} = \mathbf{G}_{gate}^\top &\mathbf{K}  \quad \quad \nabla_{\mathbf{W}_\text{down}} \mathcal{L} = \mathbf{G}_{out}^\top  \mathbf{A}  \quad \quad \nabla_{\mathbf{W}_\text{up}} \mathcal{L} = \mathbf{G}_{in}^\top \mathbf{K} 
\end{aligned}
$$
To enable efficient and stable test-time training, we adopt a momentum-based stochastic gradient descent (SGD) optimizer, augmented with input-dependent dynamic learning rates and momentum coefficients:
$$
    \begin{aligned}
\mathbf{M}_j &= \mu_j \mathbf{M}_j - \text{dot\_product}( \boldsymbol{\eta}_t,   \nabla_{\mathbf{W}} \mathcal{L}) \\
\mathbf{W}_j &= \mathbf{W}_{j-1}+ \mathbf{M}_j
\end{aligned}
$$
where $\mathbf{M}_j$  and  $\mathbf{W}_j$ denote the momentum and fast weight for the $j$-th chunk, respectively. The momentum decay coefficient is a chunk-wise, input-dependent parameter 
$\mu_j = \frac{1}{c}\sum_{t=jc+1}^{jc+c} (\text{sigmoid}\left(\text{Linear}(\boldsymbol{x}_t)\right)$ while $\boldsymbol{\eta}_t= \text{softplus}(\text{Linear}(\boldsymbol{x}_t))$ is a token-wise meta learning rate.
Under this framework, the process of optimizing the memory network via error backpropagation is referred to as the inner loop. Meanwhile, the control network responsible for generating the dynamic learning rates, momentum coefficients, and the query, key, and value vectors constitutes the outer loop.
To stabilize the internal TTT loop—particularly to mitigate gradient explosion caused by unstable meta-parameters during early training—we incorporate a carefully designed normalization mechanism. Traditional approaches such as weight normalization or gradient clipping are applied as:

$$
\boldsymbol{W} = \boldsymbol{W} \frac{\mathcal{N} }{ l^2_\text{norm}(\boldsymbol{W})} ,\quad  \nabla{\boldsymbol{W}} = \text{norm\_clip}(\nabla{\boldsymbol{W}}, \Theta)
$$
where $\mathcal{N}$ represents the $L_2$ norm of the initial memory weights (computed along the input dimension) and $\Theta$ serves as the gradient clipping threshold.
However, existing methods~\cite{zhang2025test} often apply normalization \textit{immediately after} each memory update. This ``update-then-reset'' strategy risks distorting or erasing recently learned memory structures, especially during the early stages of adaptation.
To resolve this, as illustrated in \cref{fig.pipeline}, we propose an \textit{internal normalization mechanism}, where the normalization step is inserted between the memory read and update operations. This design ensures that memory read operations are performed on unnormalized, pristine weights, while normalization is deferred until after the read is complete. This decouples memory read stability from write plasticity, enabling robust, adaptive memory learning without sacrificing the integrity of newly acquired knowledge.

For the memory module, a multi-head configuration is adopted, wherein the number of heads $n_{mem}$ is maintained in parity with the KV heads $n_{kv}$ of the Grouped-Query Attention (GQA) mechanism. A uniform dimensionality of $d=512$ is established, such that the dimension per head is defined as $d_h = d / n_{kv}$. Furthermore, non-sharing of parameters is enforced between distinct layers to preserve representational capacity. Regarding positional encoding, the memory module remains position-agnostic, whereas the SWA component utilizes Rotary Positional Embeddings (RoPE)\cite{su2023rope} to encode relative sequence order. 

In addition to the core memory architecture, the meta-learning backbone is optimized by integrating advancements from {popular} linear attention and Transformer frameworks. Specifically, one-dimensional convolutional layers are incorporated following the $QKV$ linear projections \cite{yang2024gated,gu2024mamba,behrouz2024titans} to strengthen local contextual dependencies and facilitate the contextual compression of memory content. Furthermore, distribution normalization is implemented within the $Q$ and $K$ projection branches via RMSNorm and head-wise $L_2$ normalization \cite{yang2025qwen3}, which effectively stabilizes input signal variance and accelerates network convergence. To further ensure training stability and mitigate potential attention sink phenomena \cite{qwen2025qwen3next}, an auxiliary parallel gating mechanism is also integrated into the architecture.

\subsection{Distilation pipeline} \label{sec:transformation_pipeline}

\subsubsection{Data Preprocessing}
For data preprocessing, long-sequence distillation data are sourced from \textsc{ChatQA2}~\cite{xu2024chatqa}, while short-sequence data are drawn from \textsc{ChineseInstruct}~\cite{mxode2024chineseinstruct} and \textsc{SFTv3}~\cite{junxiongdaniele2024mambainllama}. All datasets undergo a unified cleaning pipeline: format conversion, sequence length normalization, and data filtering. Specifically, all raw inputs are converted into the standardized \textsc{ChatML} format, further aligned with the Qwen3 chat template. For long-sequence data, we adopt a context length of 24K tokens; for short-sequence distillation, we set the context length to 4K tokens. Crucially, during long-sequence training, the loss is calculated only on the response tokens, discarding any extraneous prompt or instruction content. It effectively guides the memory module to focus on salient contextual information, while suppressing redundancy and noise from non-informative background structures. Consequently, enhanced memory capacity and superior generalization ability are achieved through response-only distillation.

\subsubsection{Distillation Method}
We employ an end-to-end knowledge distillation framework, where the original Transformer model serves as the teacher, guiding the training of our \textsc{AllMem} student model. The training objective is to minimize the forward Kullback-Leibler (KL) divergence between the output probability distributions of the student and teacher models. 
Prior studies~\cite{wang2024mamba,rang2025revealing} have demonstrated that incorporating a supervised cross-entropy (CE) loss can further enhance the performance of distilled models. This hybrid objective is typically expressed as:
$$
L_{total} = \alpha * L_{KL} + \beta * L_{CE}
$$
where $\alpha$ and $\beta$ are hyperparameters controlling the relative weight of the distillation loss and the supervised cross-entropy (CE) loss, respectively. We found that using only the distillation loss yields the best performance. Therefore, in our experiments, we set $\alpha=1$ and $\beta=0$.

To equip the student \textsc{AllMem} model with the ability to perform effective information compression across arbitrary sequence lengths without dependence on a fixed architecture, we adopt a randomized configuration training strategy during the knowledge distillation. In each training iteration, the number of attention sinks, the length of the sliding window, and the chunk size of memory update are independently sampled from the ranges $[0, 256]$, $[512, 8192]$, and $[512, 4096]$, respectively. Empirical results show this stochastic training policy 
{facilitates} the emergence of robust, and generalizable compression mechanisms that are invariant to specific parameter settings.

\subsubsection{On-Policy Distillation}
To further enhance the performance of the distilled student model, we adopt an \textit{on-policy distillation} strategy~\cite{yang2025qwen3,lu2025onpolicydistillation}. Specifically, the student model generates responses to instruction-based datasets, and the teacher model conducts output distribution distillation over these student-generated samples. This approach enables a fine-grained alignment between the teacher and student, further improving the student's accuracy beyond the standard end-to-end distillation framework.

We first adapted the \textsc{AllMem} model to the \texttt{vLLM} v1 inference framework~\cite{kwon2023efficient}, significantly accelerating its inference speed. To improve the efficiency of on-policy distillation, we pre-collected a dataset of responses generated by the student model on short-sequence instruction benchmarks prior to distillation. Subsequently, we conducted end-to-end distillation using this fixed dataset—this procedure is also referred to as \textit{offline on-policy distillation}~\cite{rang2025revealing}.The primary advantage of this offline approach lies in eliminating the need for real-time data generation during training, thereby reducing per-step latency and stabilizing the training pipeline.

\section{Experimental details}
We employ the Qwen3 \cite{yang2025qwen3} series models as the foundational architecture for validation. All models are trained using mixed-precision training, with gradient checkpointing and DeepSpeed ZeRO-2 enabled to optimize memory efficiency. The AdamW optimizer is used with a maximum learning rate of $1\times10^{-4}$, and the Warmup-Stable-Decay (WSD) learning rate scheduling strategy is adopted. The 0.6B model is trained on 3M tokens, while the 1.7B model is trained on 5M tokens. 

Our evaluation benchmark encompasses a diverse set of datasets across multiple dimensions: short-sequence commonsense reasoning tasks including C-Eval (5-shot) \cite{huang2023c}, ARC \cite{clark2018think}, HellaSwag \cite{zellers2019hellaswag}, and WinoGrande \cite{sakaguchi2021winogrande}; reasoning capabilities assessed via MMLU-Redux \cite{gema2025we} and GPQA-Diamond \cite{rein2024gpqa}; instruction following evaluated on IFEval \cite{zhou2023instruction}; mathematical reasoning tested on MATH-500 \cite{lightman2023let}; programming ability measured by LiveCodeBench (release v5, 2023.05–2025.01) \cite{jain2024livecodebench}; and long-context evaluation conducted on LongBench \cite{bai2024longbench}, InfiniteBench \cite{zhang2024infty}, and LV-Eval \cite{yuan2024lv}.

For all benchmarks, we follow the generation settings recommended in the Qwen3 technical report \cite{yang2025qwen3}. Specifically, for GPQA-Diamond, we perform 10 independent samplings per question and report the mean accuracy. For all other datasets, a single sampling is used. We set \texttt{max\_tokens=32768} to ensure high precision, particularly in long-context evaluation tasks.

\section{Results} \label{experiment_result}

\subsection{Performance on Short-Sequence Benchmarks}
\begin{table}[t]
\centering
\renewcommand{\arraystretch}{1.0} 
\caption{Performance comparison across diverse downstream benchmarks, including common sense reasoning, logical reasoning, instruction following, mathematical problem solving, and code generation. \textsc{C-Eval} is evaluated under a $5$-shot setting, while all other tasks are assessed in a $0$-shot setting. The \textsc{AllMem} model employs a sliding window length of $2048$ and $128$ attention sinks. \textbf{Bold} indicates the best result on each task; \underline{underlined} denotes the second-best performance.}
\label{tab:benchmark_short}
\setlength{\tabcolsep}{7pt}
\resizebox{ \textwidth}{!}{ 
\begin{tabularx}{\textwidth}{@{} c l c c c c@{}}
\toprule
& & \textbf{Qwen3-0.6B} & {\bfseries \makecell{Qwen3-0.6B\\-\textsc{AllMem}}} 
& \textbf{Qwen3-1.7B} & {\bfseries \makecell{Qwen3-1.7B\\-\textsc{AllMem}}} \\
\midrule
\multirow{5}{*}{\textit{Knowledge}} &
C-Eval & 40.6 & 42.2 & \underline{58.0} & \textbf{58.9} \\
& ARC-Easy & 70.0 & 70.2 & \underline{84.5} & \textbf{84.7} \\
& ARC-Challenge & 55.0 & 54.6 & \underline{74.2} & \textbf{74.4} \\
& HellaSwag & 40.8 & \underline{41.4} & \textbf{59.4} & \textbf{59.4} \\
& WinoGrande & 50.6 & 52.2 & \underline{53.0} & \textbf{53.9} \\
\midrule
\multirow{2}{*}{\textit{Reasoning}} &
MMLU-Redux & 44.9 & 47.05 & \underline{66.5} & \textbf{67.3} \\
& GPQA-Diamond & 24.6 & 25.1 & \textbf{28.9} & \underline{28.79}\\
\midrule
\textit{Alignment} &
$\text{IFEval}_{\text{ strict prompt}}$ & 60.6 & 58.4 & \underline{68.6} & \textbf{69.0} \\
\midrule
\textit{Math} &
MATH-500 & 48.8 & 49.8 & \underline{73.6} & \textbf{74.4}\\
\midrule
\textit{Coding} &
LiveCodeBench v5 & 13.8 & 14.3 & \textbf{25.5} & \underline{25}\\
\bottomrule
\end{tabularx}
}
\end{table}
We evaluate the performance of our conversion framework on Qwen3-0.6B and Qwen3-1.7B models. After conversion, the \textsc{AllMem} architecture introduces only $66$M additional parameters (a $9.9\%$ relative increase) on the $0.6$B model, and $126$M additional parameters ($6.8\%$ relative increase) on the $1.7$B model.

We first validate the downstream accuracy of the converted \textsc{AllMem} models. \autoref{tab:benchmark_short} presents a comparative evaluation of the \textsc{AllMem} models against their original Qwen3 counterparts across multiple downstream tasks spanning diverse categories. Results demonstrate that, across both the $0.6$B and $1.7$B model sizes, the converted \textsc{AllMem} models achieve accuracy comparable to, or even exceeding, that of the original models.
This consistent performance gain across {the two different model scales, give us confidence that our method retains the original capabilities of the teacher model.} While the extended context modeling benefits are largely attributed to the wide sliding window, maintaining high accuracy on standard short-sequence benchmarks is non-trivial, especially for alternative architectures such as Hybrid SSMs, which often suffer from performance degradation on shorter inputs with long-context extension. 

\begin{table}[t]
\centering
\scriptsize
\renewcommand{\arraystretch}{1} 
\caption{Accuracy comparison of the 0.6B \textsc{AllMem} model, full attention, sliding-window attention with attention sinks, and Mamba-enhanced memory model across long-context benchmarks: \textit{LongBench}, \textit{InfiniteBench}, and \textit{LVEval}. The sliding window length is set to 4k on \textit{LongBench} and 8k on \textit{InfiniteBench} and \textit{LVEval}, with 128 attention sinks. \textbf{Bold} indicates the best result; {\underline{underline}} indicates the second-best result.}
\label{tab:longbench-0.6b}
\setlength{\tabcolsep}{1pt}
\resizebox{ \textwidth}{!}{ 
\begin{tabular}{@{}ccccccc@{}}
\toprule
&\multicolumn{1}{c}{\bfseries \makecell{Subsets}}& \multicolumn{1}{l}{\bfseries \makecell{Avg. \\ Len.}} & \multicolumn{1}{l}{\textbf{Qwen3-0.6B}} & \multicolumn{1}{l}{\bfseries \makecell{Qwen3-0.6B\\-SWA-Sinks}} & \multicolumn{1}{l}{\bfseries \makecell{Qwen3-0.6B\\-Mamba}} & \multicolumn{1}{l}{\bfseries \makecell{Qwen3-0.6B\\-\textsc{AllMem}}}\\ \midrule
\multirow{7}{*}{\textit{LongBench}}     & dureader                & 10,656                                 & 26.03              & 23.52                         & 27.4                      & 25.71                     \\
                                        & hotpotqa                & 13,466
                                  & 27.21              & 23.67                         & 22.41                     & 28.23                    \\
                                        & musique                 & 16,363
                                 & 10.86              & 8.56                          & 7.24                      & 10.76                    \\
                                        & narrativeqa             & 29,949
                                 & 13.38              & 13.08                         & 13.53                     & 12.31                    \\
                                        & qmsum                   & 139,667
                                 & 16.86              & 16.01                         & 20.14                     & 16.32                    \\
                                        & triviaqa                & 12,208
                                 & 74.62              & 69.41                         & 71.31                     & 70.64                    \\ \cmidrule{2-7} 
                                        & {Average}        & {37052}                         & \textbf{28.16}     & {25.71}                & {27.01}            & \underline{27.33}           \\ \midrule
\multirow{2}{*}{\textit{InfiniteBench}} & longbook\_qa\_en        & \multirow{2}{*}{128k}                  & 1.76               & 3.7                           & \underline{6}                         & \textbf{8.53}                    \\
                                        & longbook\_qa\_cn        &                                        & 2.18               & 4.05                          & \underline{8.5}               & \textbf{12.54}                    \\ \midrule
\multirow{12}{*}{\textit{LV-Eval}}      & cmrc\_mixup             & \multirow{11}{*}{128k}                 & 6.31               & 5.45                          & 5.91                      & 4.89                     \\
                                        & dureader\_mixup         &                                        & 11.18              & 9.33                          & 9.25                      & 9.56                     \\
                                        & factrecall\_en          &                                        & 1.67               & 1.1                           & 1.59                      & 7.5                      \\
                                        & factrecall\_zh          &                                        & 4.49               & 2.91                          & 3.7                       & 4.8                      \\
                                        & hotpotwikiqa\_mixup     &                                        & 3.19               & 2.24                          & 3.29                      & 4.96                     \\
                                        & lic\_mixup              &                                        & 6.44               & 3.66                          & 4.02                      & 3.64                     \\
                                        & loogle\_CR\_mixup       &                                        & 5.24               & 1.7                           & 3.69                      & 3.26                     \\
                                        & loogle\_MIR\_mixup      &                                        & 1.42               & 0.88                          & 1.36                      & 1.22                     \\
                                        & loogle\_SD\_mixup       &                                        & 4.87               & 2.56                          & 2.84                      & 2.34                     \\
                                        & multifieldqa\_en\_mixup &                                        & 5.53               & 3.06                          & 4.42                      & 3.73                     \\
                                        & multifieldqa\_zh\_mixup &                                        & 6.44               & 3.06                          & 3.6                       & 3.79                     \\ \cmidrule{2-7} 
                                        & {Average}        & \multicolumn{1}{c}{{128k}}          & \textbf{5.16}   & {3.27}              & {3.97}             & \underline{4.52}     \\ \bottomrule
\end{tabular}
}
\end{table}

\subsection{Performance on Long-Sequence Benchmarks}
As shown in \autoref{tab:longbench-0.6b} and \autoref{tab:longbench-1.7b}, we further validate the effectiveness of the proposed transformation framework on long-sequence tasks across multiple long-context evaluation benchmarks: \textit{LongBench}, \textit{InfiniteBench}, and \textit{LVEval}. Similar to the approach in AHN~\cite{fang2025artificial}, we select six subtasks from \textit{LongBench} with an average sequence length of 37k for evaluation. For \textit{InfiniteBench}, the input context lengths in the Chinese and English question-answering tasks reach up to 193k and 2069k, respectively. To ensure compatibility with the Qwen3 tokenizer’s maximum sequence length constraint, we apply truncation to these inputs with 128k context length. 
In contrast, \textit{LVEval} provides subtasks with varying context lengths, allowing us to directly select the subset corresponding to a 128k context window for testing.

To evaluate the effectiveness of the memory module, we implement and compare {two} baseline methods: a model based on{} sliding-window attention {with ``sink'' tokens}, and a {hybrid model using} the linear attention architecture of Mamba. Specifically, the Mamba-enhanced baseline replaces only the {\textsc{AllMem}} module in the original structure with the Mamba-2 module, while maintaining consistent model parameters. Experimental results indicate that, on the \textit{LongBench} dataset, the transformed {\textsc{AllMem}} network achieves performance comparable to full attention mechanisms employing only a 4k local attention window when processing sequences of average length 37k. On \textit{InfiniteBench} and \textit{LVEval}, the 1.7B model achieves higher accuracy than full-attention models trained under 128k context lengths, respectively, using just an 8k local attention window.

It should be emphasized that the generation hyperparameters used in this experiment are drawn from the recommended settings of Qwen3, but have not been specifically optimized for the memory-aware model. For instance, the {\textsc{AllMem}-1.7B} model achieves an average accuracy of 32.95\% on the \textit{LongBench} subtasks using only greedy decoding--a result that outperforms the default Qwen3 generation strategy. Moreover, across both 0.6B and 1.7B parameter scales, the {\textsc{AllMem}} models consistently outperform Mamba and Sink-based sliding-window attention baselines in terms of overall performance across the three major long-context benchmarks. These findings further validate the efficacy of the {\textsc{AllMem}} architectural design and its distillation framework in handling long-sequence modeling tasks.

\begin{table}[t]
\centering
\scriptsize
\renewcommand{\arraystretch}{1.0} 
\caption{Accuracy comparison of the 1.7B \textsc{AllMem} model, full attention, sliding-window attention with attention sinks, and Mamba-enhanced memory model across long-context benchmarks: \textit{LongBench}, \textit{LVEval}, and \textit{InfiniteBench}. The sliding window length is set to 4k on \textit{LongBench} and 8k on \textit{InfiniteBench} and \textit{LVEval}, with 128 attention sinks. \textbf{Bold} indicates the best result; \underline{underline} indicates the second-best result.}
\label{tab:longbench-1.7b}
\setlength{\tabcolsep}{1pt}
\resizebox{ \textwidth}{!}{ 
\begin{tabular}{@{}ccccccc@{}}
\toprule
&\multicolumn{1}{c}{\bfseries \makecell{Subsets}}& \multicolumn{1}{l}{\bfseries \makecell{Avg. \\ Len.}} & \multicolumn{1}{l}{\textbf{Qwen3-1.7B}} & \multicolumn{1}{l}{\bfseries \makecell{Qwen3-1.7B\\-SWA-Sinks}} & \multicolumn{1}{l}{\bfseries \makecell{Qwen3-1.7B\\-Mamba}} & \multicolumn{1}{l}{\bfseries \makecell{Qwen3-1.7B\\-{\textsc{AllMem}}}}\\ \midrule

\multirow{7}{*}{\textit{Longench}}     & dureader                & 10,656
                                 & 24.28               & 24.97                         & 29.37                     & 26.62                    \\
                                        & hotpotqa                & 13,466
                                  & 38.65               & 28.69                         & 28.07                     & 31.66                    \\
                                        & musique                 & 16,363
                                 & 15.92               & 11.76                         & 8.79                      & 14.87                    \\
                                        & narrativeqa             & 29,949
                                 & 18.9                & 16.41                         & 16.94                     & 17.32                    \\
                                        & qmsum                   & 139,667
                                 & 17.76               & 16.41                         & 21.37                     & 17.15                    \\
                                        & triviaqa                & 12,208
                                  & 85.81               & 83.61                         & 84.9                      & 85.12                    \\ \cmidrule{2-7} 
                                        & Average                 & 37052                             & \textbf{33.55}               & 30.31                         & 31.57               & \underline{32.12}                    \\ \midrule
\multirow{2}{*}{\textit{InfiniteBench}} & longbook\_qa\_en        & \multirow{2}{*}{128k}                  & 2.43                & 4.16                          & \underline{5.4}                       & \textbf{7.84}                     \\
                                        & longbook\_qa\_cn       &                                        & 2.42                & 4.48                          & \underline{5.4}                       & \textbf{12.74}                    \\ \midrule
\multirow{12}{*}{\textit{LV-Eval}}      & cmrc\_mixup             & \multirow{11}{*}{128k}                 & 6.11                & 5.27                          & 7.05                      & 5.64                     \\
                                        & dureader\_mixup         &                                        & 11.97               & 9.46                          & 10.64                     & 9.91                     \\
                                        & factrecall\_en          &                                        & 1.8                 & 1                             & 1                         & 6.83                     \\
                                        & factrecall\_zh          &                                        & 0.38                & 0.31                          & 0.5                       & 7                        \\
                                        & hotpotwikiqa\_mixup     &                                        & 4.74                & 4.96                          & 2.73                      & 6.81                     \\
                                        & lic\_mixup              &                                        & 6.01                & 2.83                          & 5.11                      & 4.61                     \\
                                        & loogle\_CR\_mixup       &                                        & 3.26                & 1.71                          & 2.52                      & 2.24                     \\
                                        & loogle\_MIR\_mixup      &                                        & 1.72                & 1.34                          & 1.66                      & 1.14                     \\
                                        & loogle\_SD\_mixup       &                                        & 6.6                 & 4.61                          & 4.25                      & 4.42                     \\
                                        & multifieldqa\_en\_mixup &                                        & 7.74                & 4.77                          & 5.81                      & 5.78                     \\
                                        & multifieldqa\_zh\_mixup &                                        & 7.86                & 5.14                          & 6.72                      & 6.8                      \\ \cmidrule{2-7} 
                                        & Average                & \multicolumn{1}{l}{128k}                & \underline{5.29}                & 3.76                          & 4.36             & \textbf{5.56}                     \\ \bottomrule
\end{tabular}
}
\end{table}

\subsection{Computational Cost and Memory Overhead} 
\begin{figure}[t]
	\centering
    \includegraphics[width=\linewidth]{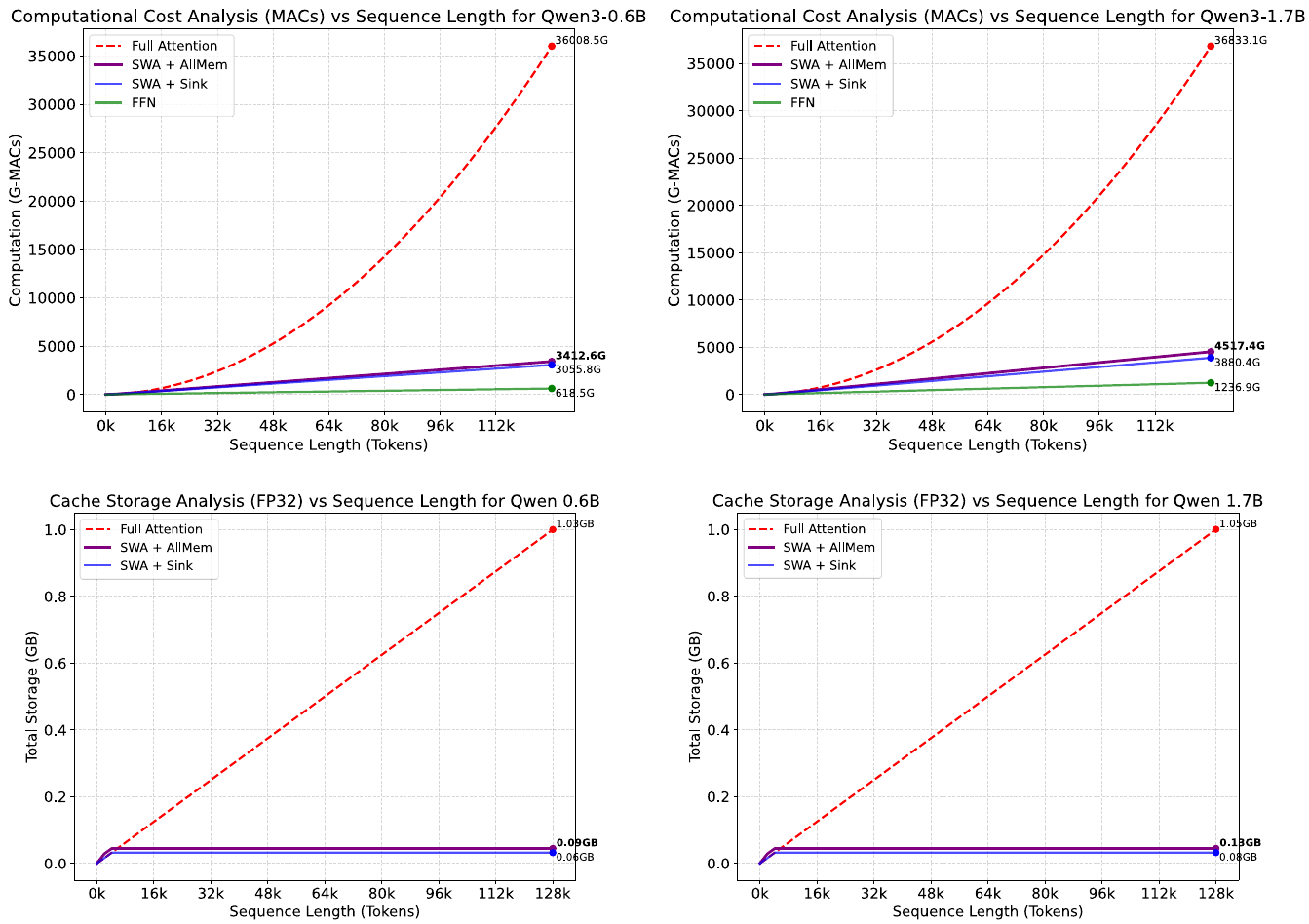}
    \caption{Comparison of FLOPs and cache size across different model sizes for {\textsc{AllMem}} memory, full attention, sink-based sliding window attention, and Channel Mixer, as sequence length increases. The sliding window size is 4k, and the update chunk size is 2k.}
	\label{fig.mac_stat}
\end{figure}

To systematically evaluate the superiority of the memory module over the full-attention mechanism in terms of computational cost and cache storage requirements, we compare{} the number of floating-point operations (FLOPs) and cache storage demands of the \textsc{AllMem} memory module{ with} full attention, and sliding-window attention with sink {tokens} across different model scales{, and sequence lengths}. In the experiments, the sliding-window size is fixed at 4096, and the update chunk size is set to 2048. As shown in \cref{fig.mac_stat}, when the context length extends to 128k, the computational cost and cache storage requirement of the \textsc{AllMem} memory model are reduced to approximately $1/9$ of those of the full-attention mechanism. Notably, the cache capacity of the \textsc{AllMem} model remains constant regardless of sequence length, whereas the cache demand of the full-attention layer grows linearly with sequence length, leading to a dramatic increase in memory overhead in long-sequence scenarios. This demonstrates that the \textsc{AllMem} architecture achieves significant advantages in computational efficiency and storage scalability for long-sequence modeling, making it well-suited for efficient sequence modeling tasks requiring ultra-long context processing.

\section{Conclusion}
In this work, we introduce \textsc{AllMem}, a long-sequence efficient LLM architecture, along with a distillation method tailored for pre-trained Transformers. This approach enables the seamless transition of classic, pre-trained decoder-only Transformers into architectures optimized for efficient long-sequence modeling. The converted models not only match or exceed the accuracy of full-attention mechanisms across both short and long benchmarks but also achieve $O(L)$ computational complexity and $O(1)$ storage complexity through the use of fixed-length sliding windows and fixed-size memory modules. This effectively resolves the scalability bottlenecks inherent in the full-attention mechanism for long sequences. 

Our findings demonstrate that employing a test-time learning {} memory architecture can effectively compensate for the precision loss typically induced by local attention, enabling the efficient compression of long-sequence information. Furthermore, we provide a comprehensive training pipeline designed for these memory-based structures, contributing to the exploration of next-generation long-sequence attention mechanisms. Notably, the proposed parameterized memory mechanism possesses the potential for deep integration with external persistent memory systems (e.g., RAG, Engram \cite{arslan2024survey,cheng2026conditional}), facilitating the construction of a multi-level memory hierarchy: a short-term perceptual memory {and a} long-term {parameterized/}persistent memory.

\bibliographystyle{unsrt}
\bibliography{ref}   

@article{vaswani2017attention,
  title={Attention is all you need},
  author={Vaswani, Ashish and Shazeer, Noam and Parmar, Niki and Uszkoreit, Jakob and Jones, Llion and Gomez, Aidan N and Kaiser, {\L}ukasz and Polosukhin, Illia},
  journal={Advances in neural information processing systems},
  volume={30},
  year={2017}
}

@article{bahdanau2014neural,
  title={Neural machine translation by jointly learning to align and translate},
  author={Bahdanau, Dzmitry and Cho, Kyunghyun and Bengio, Yoshua},
  journal={arXiv preprint arXiv:1409.0473},
  year={2014}
}

@inproceedings{gu2024mamba,
  title={Mamba: Linear-time sequence modeling with selective state spaces},
  author={Gu, Albert and Dao, Tri},
  booktitle={First conference on language modeling},
  year={2024}
}

@article{huang2023c,
  title={C-eval: A multi-level multi-discipline chinese evaluation suite for foundation models},
  author={Huang, Yuzhen and Bai, Yuzhuo and Zhu, Zhihao and Zhang, Junlei and Zhang, Jinghan and Su, Tangjun and Liu, Junteng and Lv, Chuancheng and Zhang, Yikai and Fu, Yao and others},
  journal={Advances in Neural Information Processing Systems},
  volume={36},
  pages={62991--63010},
  year={2023}
}

@article{clark2018think,
  title={Think you have solved question answering? try arc, the ai2 reasoning challenge},
  author={Clark, Peter and Cowhey, Isaac and Etzioni, Oren and Khot, Tushar and Sabharwal, Ashish and Schoenick, Carissa and Tafjord, Oyvind},
  journal={arXiv preprint arXiv:1803.05457},
  year={2018}
}

@article{zellers2019hellaswag,
  title={Hellaswag: Can a machine really finish your sentence?},
  author={Zellers, Rowan and Holtzman, Ari and Bisk, Yonatan and Farhadi, Ali and Choi, Yejin},
  journal={arXiv preprint arXiv:1905.07830},
  year={2019}
}

@article{sakaguchi2021winogrande,
  title={Winogrande: An adversarial winograd schema challenge at scale},
  author={Sakaguchi, Keisuke and Bras, Ronan Le and Bhagavatula, Chandra and Choi, Yejin},
  journal={Communications of the ACM},
  volume={64},
  number={9},
  pages={99--106},
  year={2021},
  publisher={ACM New York, NY, USA}
}

@inproceedings{gema2025we,
  title={Are we done with mmlu?},
  author={Gema, Aryo Pradipta and Leang, Joshua Ong Jun and Hong, Giwon and Devoto, Alessio and Mancino, Alberto Carlo Maria and Saxena, Rohit and He, Xuanli and Zhao, Yu and Du, Xiaotang and Madani, Mohammad Reza Ghasemi and others},
  booktitle={Proceedings of the 2025 Conference of the Nations of the Americas Chapter of the Association for Computational Linguistics: Human Language Technologies (Volume 1: Long Papers)},
  pages={5069--5096},
  year={2025}
}

@inproceedings{rein2024gpqa,
  title={Gpqa: A graduate-level google-proof q\&a benchmark},
  author={Rein, David and Hou, Betty Li and Stickland, Asa Cooper and Petty, Jackson and Pang, Richard Yuanzhe and Dirani, Julien and Michael, Julian and Bowman, Samuel R},
  booktitle={First Conference on Language Modeling},
  year={2024}
}

@article{zhou2023instruction,
  title={Instruction-following evaluation for large language models},
  author={Zhou, Jeffrey and Lu, Tianjian and Mishra, Swaroop and Brahma, Siddhartha and Basu, Sujoy and Luan, Yi and Zhou, Denny and Hou, Le},
  journal={arXiv preprint arXiv:2311.07911},
  year={2023}
}

@inproceedings{lightman2023let,
  title={Let's verify step by step},
  author={Lightman, Hunter and Kosaraju, Vineet and Burda, Yuri and Edwards, Harrison and Baker, Bowen and Lee, Teddy and Leike, Jan and Schulman, John and Sutskever, Ilya and Cobbe, Karl},
  booktitle={The Twelfth International Conference on Learning Representations},
  year={2023}
}

@article{jain2024livecodebench,
  title={Livecodebench: Holistic and contamination free evaluation of large language models for code},
  author={Jain, Naman and Han, King and Gu, Alex and Li, Wen-Ding and Yan, Fanjia and Zhang, Tianjun and Wang, Sida and Solar-Lezama, Armando and Sen, Koushik and Stoica, Ion},
  journal={arXiv preprint arXiv:2403.07974},
  year={2024}
}

@inproceedings{bai2024longbench,
    title = "{L}ong{B}ench: A Bilingual, Multitask Benchmark for Long Context Understanding",
    author = "Bai, Yushi and Lv, Xin  and Zhang, Jiajie  and Lyu, Hongchang  and
      Tang, Jiankai  and Huang, Zhidian  and Du, Zhengxiao  and Liu, Xiao  and Zeng, Aohan  and Hou, Lei  and Dong, Yuxiao  and Tang, Jie  and Li, Juanzi",
    booktitle = "Proceedings of the 62nd Annual Meeting of the Association for Computational Linguistics (Volume 1: Long Papers)",
    month = aug,
    year = "2024",
    address = "Bangkok, Thailand",
    publisher = "Association for Computational Linguistics",
    url = "https://aclanthology.org/2024.acl-long.172",
    doi = "10.18653/v1/2024.acl-long.172",
    pages = "3119--3137",
}

@article{zhang2024infty,
  title={$\infty$ Bench: Extending Long Context Evaluation Beyond 100K Tokens},
  author={Zhang, Xinrong and Chen, Yingfa and Hu, Shengding and Xu, Zihang and Chen, Junhao and Hao, Moo Khai and Han, Xu and Thai, Zhen Leng and Wang, Shuo and Liu, Zhiyuan and others},
  journal={arXiv preprint arXiv:2402.13718},
  year={2024}
}

@article{yuan2024lv,
  title={Lv-eval: A balanced long-context benchmark with 5 length levels up to 256k},
  author={Yuan, Tao and Ning, Xuefei and Zhou, Dong and Yang, Zhijie and Li, Shiyao and Zhuang, Minghui and Tan, Zheyue and Yao, Zhuyu and Lin, Dahua and Li, Boxun and others},
  journal={arXiv preprint arXiv:2402.05136},
  year={2024}
}

@article{yang2025qwen3,
  title={Qwen3 technical report},
  author={Yang, An and Li, Anfeng and Yang, Baosong and Zhang, Beichen and Hui, Binyuan and Zheng, Bo and Yu, Bowen and Gao, Chang and Huang, Chengen and Lv, Chenxu and others},
  journal={arXiv preprint arXiv:2505.09388},
  year={2025}
}

@article{wang2024mamba,
  title={The mamba in the llama: Distilling and accelerating hybrid models},
  author={Wang, Junxiong and Paliotta, Daniele and May, Avner and Rush, Alexander and Dao, Tri},
  journal={Advances in Neural Information Processing Systems},
  volume={37},
  pages={62432--62457},
  year={2024}
}

@article{rang2025revealing,
  title={Revealing the Power of Post-Training for Small Language Models via Knowledge Distillation},
  author={Rang, Miao and Bi, Zhenni and Zhou, Hang and Chen, Hanting and Xiao, An and Guo, Tianyu and Han, Kai and Chen, Xinghao and Wang, Yunhe},
  journal={arXiv preprint arXiv:2509.26497},
  year={2025}
}

@article{lu2025onpolicydistillation,
  author = {Kevin Lu and Thinking Machines Lab},
  title = {On-Policy Distillation},
  journal = {Thinking Machines Lab: Connectionism},
  year = {2025},
  note = {https://thinkingmachines.ai/blog/on-policy-distillation},
  doi = {10.64434/tml.20251026},
}

@inproceedings{kwon2023efficient,
  title={Efficient Memory Management for Large Language Model Serving with PagedAttention},
  author={Woosuk Kwon and Zhuohan Li and Siyuan Zhuang and Ying Sheng and Lianmin Zheng and Cody Hao Yu and Joseph E. Gonzalez and Hao Zhang and Ion Stoica},
  booktitle={Proceedings of the ACM SIGOPS 29th Symposium on Operating Systems Principles},
  year={2023}
}

@article{fang2025artificial,
  title={Artificial hippocampus networks for efficient long-context modeling},
  author={Fang, Yunhao and Yu, Weihao and Zhong, Shu and Ye, Qinghao and Xiong, Xuehan and Wei, Lai},
  journal={arXiv preprint arXiv:2510.07318},
  year={2025}
}

@misc{mxode2024chineseinstruct,
  author       = {Mxode},
  title        = {Chinese-Instruct: A Chinese Instruction-Tuning Dataset},
  year         = {2024},
  publisher    = {Hugging Face},
  howpublished = {\url{https://huggingface.co/datasets/Mxode/Chinese-Instruct}},
  note         = {Accessed: 2026-01-09}
}

@article{xu2024chatqa,
  title={Chatqa 2: Bridging the gap to proprietary llms in long context and rag capabilities},
  author={Xu, Peng and Ping, Wei and Wu, Xianchao and Xu, Chejian and Liu, Zihan and Shoeybi, Mohammad and Catanzaro, Bryan},
  journal={arXiv preprint arXiv:2407.14482},
  year={2024}
}

@article{junxiongdaniele2024mambainllama,
  title   = {The Mamba in the Llama: Distilling and Accelerating Hybrid Models},
  author  = {Junxiong Wang and Daniele Paliotta and Avner May and Alexander M. Rush and Tri Dao},
  journal = {arXiv preprint arXiv:2408.15237},
  year    = {2024}
}

@article{yang2024gated,
  title={Gated delta networks: Improving mamba2 with delta rule},
  author={Yang, Songlin and Kautz, Jan and Hatamizadeh, Ali},
  journal={arXiv preprint arXiv:2412.06464},
  year={2024}
}

@article{yang2023gated,
  title={Gated linear attention transformers with hardware-efficient training},
  author={Yang, Songlin and Wang, Bailin and Shen, Yikang and Panda, Rameswar and Kim, Yoon},
  journal={arXiv preprint arXiv:2312.06635},
  year={2023}
}

@article{zuo2025falcon,
  title={Falcon-h1: A family of hybrid-head language models redefining efficiency and performance},
  author={Zuo, Jingwei and Velikanov, Maksim and Chahed, Ilyas and Belkada, Younes and Rhayem, Dhia Eddine and Kunsch, Guillaume and Hacid, Hakim and Yous, Hamza and Farhat, Brahim and Khadraoui, Ibrahim and others},
  journal={arXiv preprint arXiv:2507.22448},
  year={2025}
}

@article{dong2024hymba,
  title={Hymba: A hybrid-head architecture for small language models},
  author={Dong, Xin and Fu, Yonggan and Diao, Shizhe and Byeon, Wonmin and Chen, Zijia and Mahabaleshwarkar, Ameya Sunil and Liu, Shih-Yang and Van Keirsbilck, Matthijs and Chen, Min-Hung and Suhara, Yoshi and others},
  journal={arXiv preprint arXiv:2411.13676},
  year={2024}
}

@article{behrouz2024titans,
  title={Titans: Learning to memorize at test time},
  author={Behrouz, Ali and Zhong, Peilin and Mirrokni, Vahab},
  journal={arXiv preprint arXiv:2501.00663},
  year={2024}
}

@misc{qwen2025qwen3next,
  title = {Qwen3-Next: Towards Ultimate Training \& Inference Efficiency},
  author = {Qwen Team},
  year = {2025},
  month = {September},
  day = {10},
  howpublished = {\url{https://qwen.ai/blog?id=4074cca80393150c248e508aa62983f9cb7d27cd&from=research.latest-advancements-list}},
  note = {Accessed: 2026-01-09}
}

@article{li2025minimax,
  title={Minimax-01: Scaling foundation models with lightning attention},
  author={Li, Aonian and Gong, Bangwei and Yang, Bo and Shan, Boji and Liu, Chang and Zhu, Cheng and Zhang, Chunhao and Guo, Congchao and Chen, Da and Li, Dong and others},
  journal={arXiv preprint arXiv:2501.08313},
  year={2025}
}

@article{team2025kimi,
  title={Kimi linear: An expressive, efficient attention architecture},
  author={Team, Kimi and Zhang, Yu and Lin, Zongyu and Yao, Xingcheng and Hu, Jiaxi and Meng, Fanqing and Liu, Chengyin and Men, Xin and Yang, Songlin and Li, Zhiyuan and others},
  journal={arXiv preprint arXiv:2510.26692},
  year={2025}
}

@article{agarwal2025gpt,
  title={gpt-oss-120b \& gpt-oss-20b model card},
  author={Agarwal, Sandhini and Ahmad, Lama and Ai, Jason and Altman, Sam and Applebaum, Andy and Arbus, Edwin and Arora, Rahul K and Bai, Yu and Baker, Bowen and Bao, Haiming and others},
  journal={arXiv preprint arXiv:2508.10925},
  year={2025}
}

@article{zhang2025test,
  title={Test-time training done right},
  author={Zhang, Tianyuan and Bi, Sai and Hong, Yicong and Zhang, Kai and Luan, Fujun and Yang, Songlin and Sunkavalli, Kalyan and Freeman, William T and Tan, Hao},
  journal={arXiv preprint arXiv:2505.23884},
  year={2025}
}

@article{behrouz2025nested,
  title={Nested learning: The illusion of deep learning architectures},
  author={Behrouz, Ali and Razaviyayn, Meisam and Zhong, Peilin and Mirrokni, Vahab},
  journal={arXiv preprint arXiv:2512.24695},
  year={2025}
}

@inproceedings{sun2020test,
  title={Test-time training with self-supervision for generalization under distribution shifts},
  author={Sun, Yu and Wang, Xiaolong and Liu, Zhuang and Miller, John and Efros, Alexei and Hardt, Moritz},
  booktitle={International conference on machine learning},
  pages={9229--9248},
  year={2020},
  organization={PMLR}
}

@article{dao2024transformers,
  title={Transformers are ssms: Generalized models and efficient algorithms through structured state space duality},
  author={Dao, Tri and Gu, Albert},
  journal={arXiv preprint arXiv:2405.21060},
  year={2024}
}

@book{fan2018local,
  title={Local polynomial modelling and its applications: monographs on statistics and applied probability 66},
  author={Fan, Jianqing},
  year={2018},
  publisher={Routledge}
}

@article{ji2007coordinated,
  title={Coordinated memory replay in the visual cortex and hippocampus during sleep},
  author={Ji, Daoyun and Wilson, Matthew A},
  journal={Nature neuroscience},
  volume={10},
  number={1},
  pages={100--107},
  year={2007},
  publisher={Nature Publishing Group US New York}
}

@article{kirkpatrick2017overcoming,
  title={Overcoming catastrophic forgetting in neural networks},
  author={Kirkpatrick, James and Pascanu, Razvan and Rabinowitz, Neil and Veness, Joel and Desjardins, Guillaume and Rusu, Andrei A and Milan, Kieran and Quan, John and Ramalho, Tiago and Grabska-Barwinska, Agnieszka and others},
  journal={Proceedings of the national academy of sciences},
  volume={114},
  number={13},
  pages={3521--3526},
  year={2017},
  publisher={National Academy of Sciences}
}

@article{cheng2026conditional,
  title={Conditional Memory via Scalable Lookup: A New Axis of Sparsity for Large Language Models},
  author={Cheng, Xin and Zeng, Wangding and Dai, Damai and Chen, Qinyu and Wang, Bingxuan and Xie, Zhenda and Huang, Kezhao and Yu, Xingkai and Hao, Zhewen and Li, Yukun and others},
  journal={arXiv preprint arXiv:2601.07372},
  year={2026}
}

@article{sun2024learning,
  title={Learning to (learn at test time): Rnns with expressive hidden states},
  author={Sun, Yu and Li, Xinhao and Dalal, Karan and Xu, Jiarui and Vikram, Arjun and Zhang, Genghan and Dubois, Yann and Chen, Xinlei and Wang, Xiaolong and Koyejo, Sanmi and others},
  journal={arXiv preprint arXiv:2407.04620},
  year={2024}
}

@article{xiao2023efficient,
  title={Efficient streaming language models with attention sinks},
  author={Xiao, Guangxuan and Tian, Yuandong and Chen, Beidi and Han, Song and Lewis, Mike},
  journal={arXiv preprint arXiv:2309.17453},
  year={2023}
}

@article{lieber2024jamba,
  title={Jamba: A hybrid transformer-mamba language model},
  author={Lieber, Opher and Lenz, Barak and Bata, Hofit and Cohen, Gal and Osin, Jhonathan and Dalmedigos, Itay and Safahi, Erez and Meirom, Shaked and Belinkov, Yonatan and Shalev-Shwartz, Shai and others},
  journal={arXiv preprint arXiv:2403.19887},
  year={2024}
}

@article{gu2025jet,
  title={Jet-nemotron: Efficient language model with post neural architecture search},
  author={Gu, Yuxian and Hu, Qinghao and Yang, Shang and Xi, Haocheng and Chen, Junyu and Han, Song and Cai, Han},
  journal={arXiv preprint arXiv:2508.15884},
  year={2025}
}

@article{arslan2024survey,
  title={A Survey on RAG with LLMs},
  author={Arslan, Muhammad and Ghanem, Hussam and Munawar, Saba and Cruz, Christophe},
  journal={Procedia computer science},
  volume={246},
  pages={3781--3790},
  year={2024},
  publisher={Elsevier}
}

@misc{vonoswald2025mesanetsequencemodelinglocally,
      title={MesaNet: Sequence Modeling by Locally Optimal Test-Time Training}, 
      author={Johannes von Oswald and Nino Scherrer and Seijin Kobayashi and Luca Versari and Songlin Yang and Maximilian Schlegel and Kaitlin Maile and Yanick Schimpf and Oliver Sieberling and Alexander Meulemans and Rif A. Saurous and Guillaume Lajoie and Charlotte Frenkel and Razvan Pascanu and Blaise Agüera y Arcas and João Sacramento},
      year={2025},
      eprint={2506.05233},
      archivePrefix={arXiv},
      primaryClass={cs.LG},
      url={https://arxiv.org/abs/2506.05233}, 
}

@misc{su2023rope,
      title={RoFormer: Enhanced Transformer with Rotary Position Embedding}, 
      author={Jianlin Su and Yu Lu and Shengfeng Pan and Ahmed Murtadha and Bo Wen and Yunfeng Liu},
      year={2023},
      eprint={2104.09864},
      archivePrefix={arXiv},
      primaryClass={cs.CL},
      url={https://arxiv.org/abs/2104.09864}, 
}

@misc{qwen3next,
	author = {},
	title = {{Q}wen --- qwen.ai},
	howpublished = {\url{https://qwen.ai/blog?id=4074cca80393150c248e508aa62983f9cb7d27cd\&from=research.latest-advancements-list}},
	year = {},
	note = {[Accessed 14-02-2026]},
}

@misc{hfRing251T,
	author = {},
	title = {inclusion{A}{I}/{R}ing-2.5-1{T} · {H}ugging {F}ace --- huggingface.co},
	howpublished = {\url{https://huggingface.co/inclusion{A}{I}/{R}ing-2.5-1{T}}},
	year = {},
	note = {[Accessed 14-02-2026]},
}

\end{document}